\documentclass[letterpaper, 10 pt, conference]{ieeeconf}  

\IEEEoverridecommandlockouts                              

\usepackage{graphicx} 
\usepackage{amsmath} 
\usepackage{hyperref}
\usepackage{float}
\usepackage{color}
\usepackage[font=footnotesize,skip=2pt]{caption}

\title{\LARGE \bf
	Task Planning for Mobile Manipulation in Retail Stores using Foundation Models with Iterative Re-planning \vspace{-1em}
}

\author{Vismay Vakharia$^{*}$, Sanjana Garai, Rolif Lima, Nijil George, Vighnesh Vatsal, Kaushik Das
	\thanks{The authors are with TCS Research, Tata Consultancy Services Ltd., Bengaluru - 560066, Karnataka, India. $^{*}$Corresponding author, e-mail:
		{\tt\small vismay.vakharia@tcs.com}}
}

\begin{document}
	
	\maketitle
	\thispagestyle{empty}
	\pagestyle{empty}

	\begin{abstract}
		Automation in industries such as retail, warehousing and logistics presents opportunities for greater throughput, cost reduction and mitigation of disruptions from labour shortages. Previously, such efforts have focused on back-room operations involving packing and sorting in relatively structured environments. With advances in robotic mobile manipulation hardware and foundation models, automation can now be applied to more variable and human-centric environments such as retail store shelves. In this work, we present a task-planning approach using Large Language Models (LLMs) and Vision-Language Models (VLMs) to address the restocking problem in retail scenarios such as supermarkets. We demonstrate this system on a custom omnidirectional mobile manipulation platform, with user-driven prompts and a feedback-based iterative re-planning approach for error correction. The end-to-end system is validated in a PyBullet simulation environment for pick-and-place tasks.   
	\end{abstract}

	\section{INTRODUCTION}

	The modern retail industry is rapidly adopting robotics-enabled solutions, driven by the competitive nature of the sector and labour shortages, particularly in developed countries. Automating a retail store using mobile robots presents significant challenges, as evidenced by international robotics competitions such as the "Amazon Picking Challenge"~\cite{correll2016analysis} and "Future Convenience Store Challenge"~\cite{wada2017new}. The robots tasked with order picking, restocking, and organizing must decompose high-level goals into sequenced sub-tasks and generate efficient motion plans -- all while navigating real-world uncertainty~\cite{guha2022robots}.
	
	Traditionally, this has been achieved using Task and Motion Planning (TAMP) \cite{haslum2024universal, kattepur2020roboplanner} frameworks, that rely on symbolic reasoning that is manually integrated with continuous motion control. However, these methods are often domain-specific and rigid in their operation.
	
	Recent work shows that Large Language Models (LLMs) can transform TAMP by replacing rigid rule-based systems with flexible, general-purpose reasoning~\cite{gupta2024action, chen2024autotamp}. Pretrained on vast internet-scale text corpora using masked language modelling and autoregressive prediction objectives, LLMs can infer structured task sequences without the need for fine-tuning~\cite{huang2022language}, adapt to environmental context, and even recover from failures without domain-specific engineering or retraining~\cite{bhat2024grounding}.
	
	In this study, we explore this new paradigm using a custom-built, omnidirectional, dual-arm mobile manipulator~\cite{smc24_autonomous, smc24_shared}. The system is tasked with `order picking' in a simulated retail environment, where it must retrieve items from a given list in their respective quantities efficiently based on store layout. Our framework combines an LLM for high-level logical reasoning with a VLM for spatial understanding, using execution feedback to continually refine its plans. The proposed framework is tested in a PyBullet simulation environment.

	\section{METHOD} 
	
	The proposed framework is initialized by the user inputting the query in natural language (English) to the LLM. Along with the query, a planogram containing the description of the environment and robot-specific information consisting of the feasible symbolic actions and their respective parameter sets (table~\ref{tab:actions}) are provided to the LLM.  
	
	
	\begin{figure}
		\centering
		\includegraphics[width=0.48\textwidth]{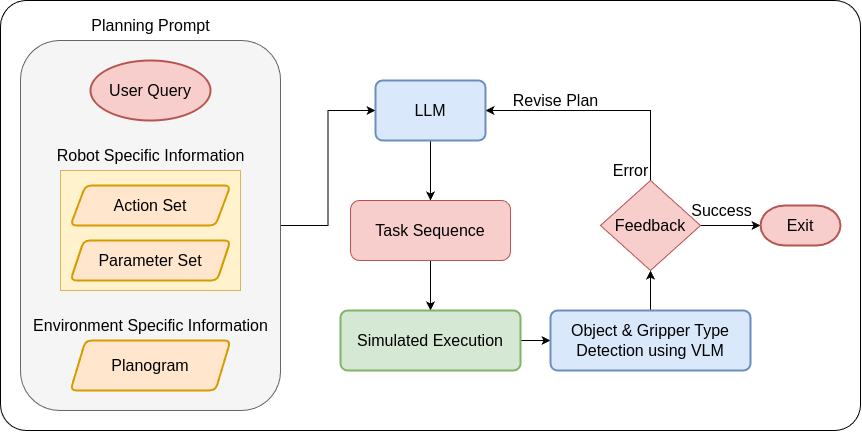}
		\caption{Framework Architecture}
		\label{fig:architecture}
		\vspace{-2em}
	\end{figure}
	
	
	
	The first stage of real-world task execution involves creating a plan that outlines a sequence of actions. The LLM is prompted to generate a task sequence from the robot's feasible actions to ensure that all the tasks are within the scope of the robot's capabilities while maintaining the context of the environment through the planogram. Each action takes specific parameters and performs the motion planning necessary to complete the task. Table \ref{tab:actions} explains the robot's feasible actions and their respective parameters.
	
	\begin{table*}[t]
		\centering
		\renewcommand{\arraystretch}{1.2}
		\begin{tabular}{|l|p{0.6\textwidth}|}
			\hline
			\textbf{Action / Parameters} & \textbf{Description} \\
			\hline  
			\texttt{navigate(table\_location)} & moves the robot base to the target \texttt{table\_location} \\
			\hline
			\texttt{scan(object\_type)} & captures an image and uses VLM to scan the environment; determines the object of \texttt{object\_type} to grasp, the gripper type and grasp orientation \\
			\hline
			\texttt{pick(object)} & plans a trajectory, grasp the \texttt{object} and places it on the tray \\
			\hline
			\texttt{place(object)} & grasps the \texttt{object} from the tray and places on the drop table \\
			\hline
		\end{tabular}
		\caption{Robot Feasible Actions and Parameters}
		\label{tab:actions}
		\vspace{-2em}
	\end{table*}
	
	The omnidirectional mobile base is equipped with an autonomous navigation system including localization and path planning algorithms. The \texttt{navigate(table\_location)} is a wrapper that interfaces the parametrized symbolic functions with a high-level motion planner that takes care of the navigation of the mobile base. A call of \texttt{navigate} makes the robot navigate to the desired target \texttt{table\_location}.
	
	In the \texttt{scan} function, the robot uses its head-mounted camera to capture an image and prompts VLM to find a \texttt{object\_type} in the robot's workspace. The VLM's spatial reasoning comes in handy to identify which object would be ideal to grasp if multiple options are found and to determine the appropriate gripper type (rigid gripper or soft gripper) as well as the end-effector grasp approach depending on how tightly packed the objects are.
	
	The \texttt{pick} and \texttt{place} functions interface with motion planners for the two UR5e arms. Once the object, gripper, and approach are selected, the planner solves inverse kinematics and computes a trajectory. In \texttt{pick}, the object is retrieved from a shelf and placed on the robot’s tray; in \texttt{place}, it's picked from the tray and placed in an appropriate slot identified using the VLM.
	
	
	While the language models are excellent at generating task plans, they're not immune to mistakes. Our framework includes real-time checks to catch errors during execution and gives immediate feedback to the LLM/VLM for on-the-fly adjustments to the plan. This iterative process ensures smooth operation without any interruptions or manual interventions.
	
	Since LLMs are inherently stochastic, they might generate invalid actions or incorrect parameters. While obvious errors—like syntax mistakes or out-of-bound values—are easy to catch due to the robot's limited and fixed capabilities, logical inconsistencies are trickier. For example, if the plan tells the robot to \texttt{navigate} to a table that doesn’t match the target object, the syntax and parameters may seem correct, but the logic is flawed. Here we utilize the VLM to cross-check the objects in view against the plan and flag the error, triggering feedback to revise the plan. Similarly, we also use this feedback loop to identify and correct missing \texttt{navigation} steps before \texttt{pick} or \texttt{place} actions. Throughout, a buffer maintaining the current execution state is maintained and included in the prompt during plan revision provided to the LLM to refine the plan on the fly, without starting from scratch.
	
	\section{SIMULATION \& EXPERIMENTS}
	
	\subsection{Simulation Setup}\label{sec:simulation}
	
	Retail store environments commonly have flat, smooth floors and narrow aisles so we use a robotic system consisting of two 6-DOF UR5e manipulators mounted on an in-house-built omnidirectional mobile base, allowing motion in all directions and in-place rotation, making it suitable for tight retail spaces. Arms are equipped with two different types of grippers, a standard 2-finger gripper for grasping rigid objects and a reconfigurable 3-finger soft gripper to manipulate deformable objects \cite{grasp_planning}.
	
	For planning, we use Mixtral AI's \texttt{Mixtral 8x22b}~\cite{jiang2024mixtral} LLM model (141B parameters, 64k context window) and \texttt{Pixtral 12b}~\cite{agrawal2024pixtral} VLM (400M vision encoder, 12B parameters multi-modal decoder and 128k context window). Both of these models are chosen for their scale, performance, and permissive open-source license.
	
	We consider an order-picking task in a simulated retail setting using PyBullet~\cite{coumans2021}. A mobile manipulator navigates the environment to pick and deliver items in specified quantities. The setup includes 8 object types placed on 4 tables arranged in a 2x2 grid with space for robot movement. The objects have three different shapes - cuboid, cylindrical and spherical. Cuboid objects represent solid, non-deformable items that can be grasped using a standard 2-finger rigid gripper. In contrast, cylindrical and spherical objects represent irregularly shaped or deformable objects that require a 3-finger soft gripper. The user query consists of any combination of these items with any count (up to 15).
	
	
	\subsection{Evaluation}
	
	To evaluate our approach, we designed a series of experiments that progressively test reliability, reasoning, and ability to recover from failure.
	
	We begin with simple, feasible order-picking tasks with shuffled item types and quantities to confirm that the system generates correct and consistent plans. Next, we test the LLM’s robustness by providing infeasible or irrelevant queries such as "play music" or "explore space" which fall well outside the robot's action space. This helps assess whether the model can correctly reject or ignore nonsensical instructions. We also craft intentionally misleading prompts to test fine-grained understanding. For instance, if the environment contains red cubes and white spheres, we include queries for items that don’t exist such as red spheres or white cubes. These tests examine whether the LLM-VLM system can accurately ground object descriptions in the observed environment. To evaluate re-planning capability, we inject errors into valid plans. For example, incorrect actions, object mismatches, wrong action sequences, or invalid parameters. The system begins execution with these flawed plans and is expected to detect the resulting failure and self-correct through feedback-driven refinement iterations.
	
	The results of these experiments demonstrate that our method enabled the robot to successfully understand the user query and perform generated tasks despite the uncertainties and challenges in the environment. The successful execution of these experiments validates the practicality and robustness of our approach, showcasing its potential for various real-world applications such as order picking, restocking, and organizing in the retail setup.
	
	\section{CONCLUSIONS}
	
	We present a task-planning framework that combines a pre-trained LLM \& VLM to generate and refine action plans without prior domain knowledge. The LLM proposes action sequences and parameters, while feedback from execution failures enables iterative re-planning. Simulated experiments show that, despite requiring multiple refinements, the system effectively handles order-packing tasks and demonstrates strong potential for retail restocking applications.
	
	
	
	\newpage
	
	\bibliographystyle{IEEEtran}
	\bibliography{references}
	
	

\end{document}